\pdfoutput=1

\documentclass[11pt]{article}

\usepackage[]{EMNLP2022}

\usepackage{times}
\usepackage{latexsym}

\usepackage[T1]{fontenc}

\usepackage[utf8]{inputenc}

\usepackage{microtype}

\usepackage{inconsolata}

\usepackage{url}
\usepackage{graphicx}  
\usepackage{url}
\usepackage{multirow}
\usepackage{comment}
\usepackage{booktabs}
\usepackage{tikz}
\usepackage[normalem]{ulem}
\usepackage[xcolor,framemethod=tikz]{mdframed}
\usepackage{amssymb}
\usepackage{subcaption}
\usepackage[shortlabels]{enumitem} 
\usepackage{bbold}
\usepackage{colortbl,array,xcolor}
\usepackage{listings}
\usepackage{color}

\definecolor{dkgreen}{rgb}{0,0.6,0}
\definecolor{gray}{rgb}{0.5,0.5,0.5}
\definecolor{mauve}{rgb}{0.58,0,0.82}
\definecolor{gray}{rgb}{0.4,0.4,0.4}
\definecolor{darkblue}{rgb}{0.0,0.0,0.6}
\definecolor{lightblue}{rgb}{0.0,0.0,0.9}
\definecolor{cyan}{rgb}{0.0,0.6,0.6}
\definecolor{darkred}{rgb}{0.6,0.0,0.0}

\lstdefinelanguage{XML}
{
  morestring=[s][\color{mauve}]{"}{"},
  morestring=[s][\color{black}]{>}{<},
  morecomment=[s]{<?}{?>},
  morecomment=[s][\color{dkgreen}]{<!--}{-->},
  stringstyle=\color{black},
  identifierstyle=\color{lightblue},
  keywordstyle=\color{blue},
  morekeywords={xmlns,xsi,noNamespaceSchemaLocation,type,id,x,y,source,target,version,tool,transRef,roleRef,objective,eventually}
}

\newcommand\corpus{\texttt{TETRA}}
\newcommand\meta{\texttt{IRC}}
\newcommand\gpt{\textbb{GPT-2}}
\newcommand\bert{\textbb{BERT}}

\newcommand\src{\textit{src}}
\newcommand\tgt{\textit{tgt}}
\definecolor{Gray}{gray}{0.935}
\newcolumntype{g}{>{\columncolor{Gray}}c}

\title{Towards Automated Document Revision: \\ Grammatical Error Correction, Fluency Edits, and Beyond}

\newcommand\riken{1}
\newcommand\allenai{2}
\newcommand\esp{3}
\newcommand\octan{4}
\newcommand\tu{5}

\author{
  Masato Mita$^{\riken}$
  Keisuke Sakaguchi$^{\allenai}$
  Masato Hagiwara$^{{\esp},{\octan}}$ \\
  \textbf{Tomoya Mizumoto}$^{\riken}$
  \textbf{Jun Suzuki}$^{{\tu},{\riken}}$ 
  \textbf{Kentaro Inui}$^{{\tu},{\riken}}$\\
  $^{\riken}$RIKEN  
  $^{\allenai}$Allen Institute for AI
  $^{\esp}$Earth Species Project\\
  $^{\octan}$Octanove Labs 
  $^{\tu}$Tohoku University
}

\begin{document}
\maketitle

\begin{abstract}
Natural language processing technology has rapidly improved automated grammatical error correction tasks, and the community begins to explore \textit{document-level} revision as one of the next challenges.
To go beyond \textit{sentence-level} automated grammatical error correction to NLP-based \textit{document-level} revision assistant, there are two major obstacles: (1) there are few public corpora with document-level revisions being annotated by professional editors, and (2) it is not feasible to elicit all possible references and evaluate the quality of revision with such references because there are infinite possibilities of revision.
This paper tackles these challenges.
First, we introduce a new document-revision corpus, \corpus, where professional editors revised academic papers sampled from the ACL anthology which contain few trivial grammatical errors that enable us to focus more on document- and paragraph-level edits such as coherence and consistency. 
Second, we explore reference-less and interpretable methods for meta-evaluation that can detect quality improvements by document revision. 
We show the uniqueness of~\corpus~compared with existing document revision corpora and demonstrate that a fine-tuned pre-trained language model can discriminate the quality of documents after revision even when the difference is subtle.
This promising result will encourage the community to further explore automated document revision models and metrics in future.
\end{abstract}

\begin{figure*}[t]
 \centering
  \includegraphics[width=0.9\linewidth]{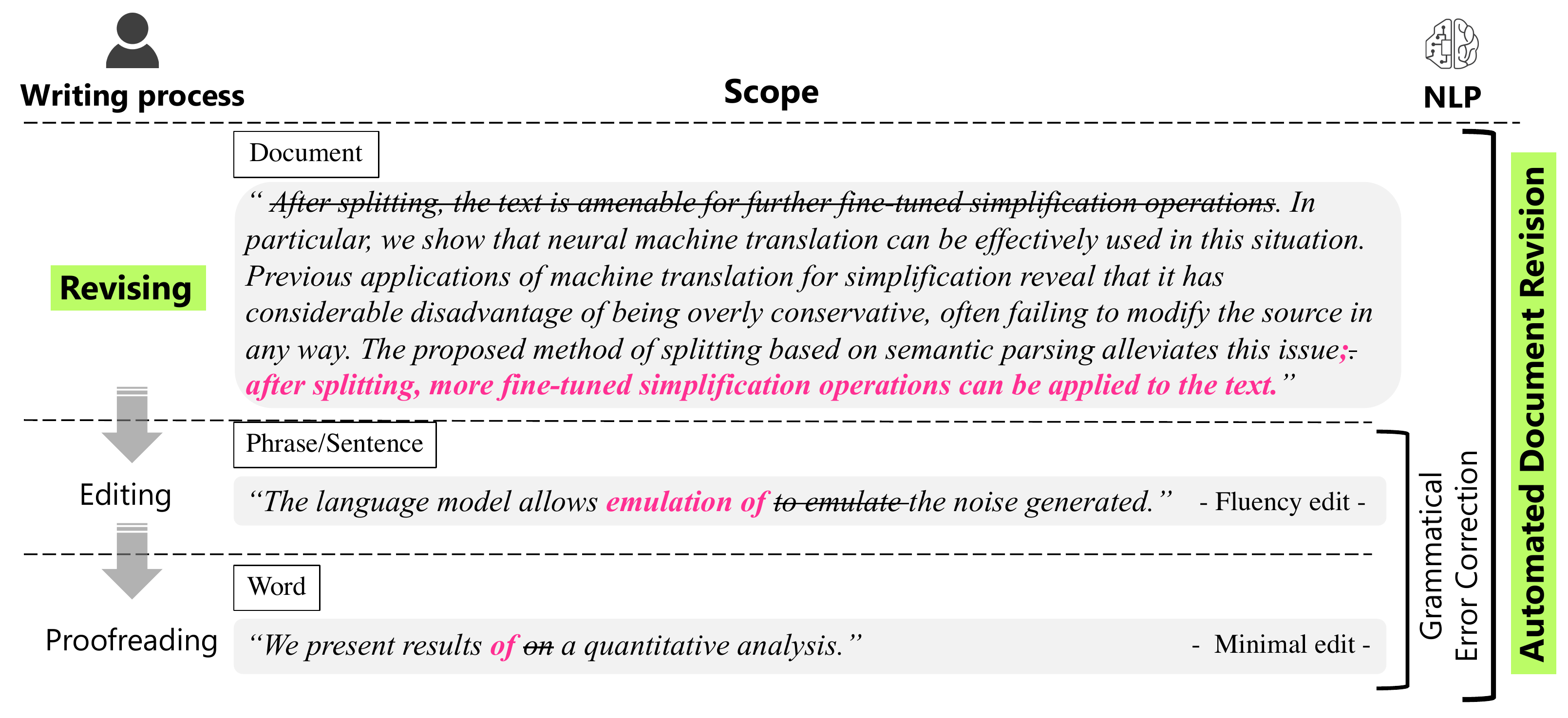}
 \caption{Overview of the scope for automated document revision. Each example is taken from~\corpus~corpus. We focus document \colorbox{lime}{revision} process which has been overlooked by grammatical error correction (GEC). Automated document revision extends the scope of GEC.}
 \label{fig:overview}
\end{figure*}

\label{sec:intro}
Document revision is a crucial process in essay and argumentative writing.
According to previous research on argumentative writing \cite{flower1981cognitive,beason1993feedback,buchman2000power,seow_2002,allal2004revision}, a typical writing process consists of three stages; \textit{revising} is the initial editing step to plan and build the overall structure of the document at a high level, \textit{editing} focuses more on sentence- or phrase-level expressions, and \textit{proofreading} checks the details such as spelling and grammatical errors (Figure~\ref{fig:overview}, left).
Although the order of the steps are not strictly determined, the typical writing process starts from a broad and high-level perspective and then narrow down the scope of edits.

Contrary to the typical human writing process, automated grammatical error correction (GEC) research in NLP initially focused on a fine-grained scope such as spelling errors~\cite{brill-moore-2000-improved,toutanova-moore-2002-pronunciation,islam-inkpen-2009-real}, closed-class parts-of-speech such as prepositions and determiners~\cite{Han:06:Journal,Nagata:06:COLING,Felice:08:COLING}. Later the research community expanded its scope to phrase- and sentence-level edits with fluency being considered~\cite{Sakaguchi_TACL16,Napoles:17:EACL} (Figure~\ref{fig:overview}, right).
However, much less work has been done on \textit{document-level} revision because of two major challenges.
First, document revisions have a broader scope (e.g., coherence and flow) than conventional GEC and fluency correction, and thus there are few publicly available corpora that have such annotations by experts (professional editors).
Second, it is challenging to evaluate the quality of revisions based on a limited number of references because there are numerous ways to revise a single document. 
This indicates that \textit{reference-less} evaluation metrics~\cite{Napoles2016,choshen-abend-2018-reference,islam-magnani-2021-end} will be suitable for assessing automated document revision models.

We tackle these challenges toward a automated \textit{document} revision, introducing a new corpus (\S\ref{sec:dataset}) and exploring possibilities for transparent evaluation methods which do not depend on gold references (\S\ref{sec:meta_eval_framwork}). 
Our new corpus, \textbf{Te}x\textbf{t}~\textbf{R}evision of \textbf{A}CL papers (\corpus)\footnote{\url{https://github.com/chemicaltree/tetra}}, consists of document-level revisions for the articles published at ACL-related venues, and is designed based on an annotation scheme that can handle edit types beyond sentences (such as argument flow) in addition to conventional word- and phrase-level edit types.
We show that \corpus~has advantages over existing corpora for document revision~\cite{lee-webster-2012-corpus,zhang-etal-2017-corpus,kashefi2022argrewrite}.
We then present a simple (meta-)evaluation method, \textbf{i}nstance-based \textbf{r}evision \textbf{c}lassification (\meta), which measures and compares the performance of evaluation metric candidates based on their accuracy for classifying which one of the given pair of snippets is a revised one.
The accuracy for \meta~is calculated per each edit type, which provides us transparent and interpretable analyses for better designing evaluation metrics in future.
We note that our contribution in this paper is not to propose a specific model or metric for automated document revision but rather present a meta-evaluation method to help measuring improvements of such models and metrics that our community will develop in future.

With the \corpus~corpus and \meta, we conduct experiments to explore whether pre-trained language models can be a good baseline metric to discriminate the original and revised snippets. 
We compared BERT~\cite{devlin-etal-2019-bert} and GPT-2~\cite{radford2019language} as baseline methods with and without fine-tuning.
The results show that the supervised method is able to choose better snippets with an accuracy of 0.85 - 0.96, indicating the feasibility of evaluation for automated document revision.

We hope that \corpus~and \meta~encourage the community to further study automated document revision modeling and metrics beyond GEC and fluency edits.

\section{Background}
The field of grammatical error correction (GEC), which has a multi-decade history, started out with the goal of detecting and correcting targeted error types and providing feedback to ESL (English as a Second Language) learners.\footnote{In this paper, we focus on GEC research literature after 2000's when statistical methods began to be applied widely. For the full history of GEC in 80's and 90's such as rule-based approaches, please refer to \citet{leacock2014automated}.} 
Initial GEC systems focused only on a small number of closed-class error types such as articles~\cite{Han:06:Journal} and prepositions~\cite{chodorow-etal-2007-detection,tetreault-chodorow-2008-ups,tetreault-etal-2010-using,cahill-etal-2013-robust,nagata-etal-2014-correcting}. 
The scope of GEC was later expanded to include errors of all types, not only closed-class words, but also verb forms, subject-verb agreement, and word choice errors~\cite{lee-seneff-2008-correcting,Tajiri:12:ACL, rozovskaya-roth-2014-building}.
This line of work resulted in establishing shared benchmark tasks~\cite{HOO2011,HOO2012,Ng:13:CoNLLST,Ng:14:CoNLLST}.

The scope of GEC has been further expanded from word-level closed-class edits to phrase- and sentence-level \textit{fluency} edits~\cite{Sakaguchi_TACL16} motivated by the observation that 
error-coded local edits do not always make the result sound natural to native speakers. 
Along with this expansion, the community has proposed new benchmark datasets~\cite{Napoles:17:EACL,bryant-etal-2019-bea,napoles-etal-2019-enabling,flachs-etal-2020-grammatical} and evaluation metrics~\cite{dahlmeier:2012:M2,felice-briscoe-2015-towards,Napoles:15:ACL,bryant:2017:automatic,napoles-etal-2019-enabling,gotou-etal-2020-taking} for sentence-to-sentence GEC, and also developed GEC models with deep neural network (DNN) approaches, finding that these models are robust against word- and phrase-level local edits in a given sentence and show human-parity performance in some benchmark datasets~\cite{yuan-briscoe-2016-grammatical,ji:2017:nested,Chollampatt:18:AAAI,ge2018reaching,kiyono-etal-2019-empirical,kaneko-etal-2020-encoder,rothe-etal-2021-simple}.
More recently,~\citet{yuan-bryant-2021-document} extend DNN models by taking longer context (e.g., previous sentence) and show improvements on sentence-level error correction (e.g., correcting verb tense).

Contrary to the rapid progress of grammar and fluency correction, few research studies have investigated revisions for document-level \textit{argumentative writing}, which requires more human effort to create corpora/datasets to start with. 
\citet{lee-webster-2012-corpus} is an initial attempt to build a document-revision corpus, which consists of 13,000 student writings with feedback comments from TESOL (Teaching English to Speakers of Other Languages) program tutors.
Although authors prepared labels for paragraph-level revisions (e.g., coherence), only 3\% of the entire revisions are annotated as paragraph-level revisions while 90\% are word-level and 7\% are sentence-level.
This is because the corpus consists of language learners' writing and the vast majority of errors are of simple grammar and fluency. 
This tells an important lesson—a corpus for document-level revision should be based on documents with grammatical and fluency edits already addressed to some extent.
In addition, this corpus is not publicly available because of the copyright. We believe that the data source for a document-level corpus should be more accessible under an open license to encourage  community-based open research in the future.

Another line of work~\cite{zhang-litman-2014-sentence,zhang-litman-2015-annotation,zhang-etal-2016-argrewrite,zhang-etal-2017-corpus,kashefi2022argrewrite} has built the ArgRewrite corpus, a collection of 86 argumentative essays each of which consists of three drafts (two cycles of revisions) with edit type labels.
In ArgRewrite corpus (both v1 and v2), nearly half of the entire edits is annotated as surface-level correction (i.e., conventional GEC or fluency edits) and the other half is annotated as content-level document revision. 
ArgRewrite corpus has advantages over~\citet{lee-webster-2012-corpus} in terms of the amount of document-level revision, but all essays are written on the same single topic. 
The topic for the first version~\cite{zhang-etal-2017-corpus} is about arguing \textit{whether the proliferation of electronic enriches or hinders the development of interpersonal relationships}, and the later version~\cite{kashefi2022argrewrite} is about arguing \textit{support or against self-driving cars}.
This limitation of topic diversity would cause an overfitting~\cite{mita-etal-2019-cross}, when it comes to developing and evaluating automated document revision models.

\section{Automated Document Revision}

\begin{table*}[t]
\centering
\small
\begin{tabular}{lllcr}
\toprule
Aspects        & Edit types (abr.)        & Definition                                        & Scope  &  \% \\ \midrule
Grammaticality & grammar, capitalization  & edits that aimed to fix spelling/grammar mistakes & S   & 19.4 \\
Fluency        & word choice, word order  & edits that aimed to increase sentence fluency     & S   & 23.7 \\ \midrule \rowcolor{Gray}
Clarity        & clarity                  &  edits that aimed to amplify meaning for clarity                                            & S/D  & 19.4 \\ \rowcolor{Gray}
Style          & style, tone              & edits that aimed to adapt the style               & S/D   &  8.0 \\ \rowcolor{Gray}
Readability    & readability &   edits that aimed to improve readability                                               & S/D & 16.8  \\ \rowcolor{Gray}
Redundancy     & redundancy, conciseness  &  edits that aimed to reduce redundancy                              & S/D   & 7.2  \\ \rowcolor{Gray}
Consistency    & consistency, flow        & edits that aimed to increase paragraph fluency    & D  & 5.5  \\  \bottomrule
\end{tabular}
\caption{Definition of edit types. S and D (in the \textit{scope} column) indicate the sentence and document, respectively. We highlight \colorbox{Gray}{edit types} that relies on beyond sentence-level context to edit.}\label{tab:definition}
\label{tab:edit_type}
\end{table*}

\newcommand{\reviewstrike}[1]{\sout{#1}}
\newcommand{\mycbox}[1]{\tikz{\path[draw=#1,fill=#1] (0,0) rectangle (0.25cm,0.25cm);}}

\definecolor{gec}{rgb}{0.9, 0.75, 0.75} 
\definecolor{flu}{rgb}{0.75, 0.75, 0.9} 
\definecolor{style}{rgb}{0.8086, 0.9141, 0.7930} 
\definecolor{cla}{rgb}{0.9727, 0.9531, 0.7617} 
\definecolor{red}{rgb}{0.9766, 0.8906, 0.8008} 
\definecolor{redund}{rgb}{0.961, 0.848, 0.902} 
\definecolor{cons}{rgb}{0.8398, 0.9219, 0.9688}  

\begin{table*}[t]
\small
\centering
\begin{center}
\begin{tabular}{p{1.0\textwidth}}
\toprule
\mycbox{gec} Grammaticality \qquad
\mycbox{flu} Fluency \qquad
\mycbox{cla} Clarity  \qquad
\mycbox{style} Style \qquad
\mycbox{red} Readability \qquad
\mycbox{redund} Redundancy \qquad
\mycbox{cons} Consistency
\\ \midrule 
This paper presents empirical studies and closely corresponding theoretical models of \setlength{\fboxsep}{0pt}\colorbox{flu}{a chart parser’s performance while}\reviewstrike{the performance of a chart parser} exhaustively parsing the Penn Treebank with the Treebank’s own \setlength{\fboxsep}{0pt}\colorbox{redund}{context-free grammar (CFG)}\reviewstrike{CFG grammar}. We show how performance is dramatically affected by rule representation and tree transformations, but little by top-down vs. bottom-up strategies. We discuss grammatical saturation\setlength{\fboxsep}{0pt}\colorbox{red}{, provide an}\reviewstrike{, including} analysis of the strongly connected components of the phrasal nonterminals in the Treebank, and model how, as sentence length increases, \setlength{\fboxsep}{0pt}\colorbox{red}{regions of the grammar are unlocked, increasing} the effective grammar rule size \reviewstrike{increases as regions of the grammar are unlocked,} \setlength{\fboxsep}{0pt}\colorbox{red}{and} yielding super-cubic observed time behavior in some configurations.
\\ \midrule
We expect this approach to yield the following three improvements. Taking advantage of the representation learned by the English model will lead to shorter training times compared to training from scratch.
Relatedly, the model trained using transfer learning \setlength{\fboxsep}{0pt}\colorbox{cons}{will require} \reviewstrike{requires} less data for an equivalent score than a German-only model. Finally, the more layers we freeze the fewer layers we \setlength{\fboxsep}{0pt}\colorbox{cons}{will} need to back-propagate through during training\setlength{\fboxsep}{0pt}\colorbox{cla}{; thus,}\reviewstrike{. Thus} we expect to see a decrease in GPU memory usage since we do not have to maintain gradients for all layers.
\\ \midrule
We present \setlength{\fboxsep}{0pt}\colorbox{gec}{the} results \setlength{\fboxsep}{0pt}\colorbox{gec}{off} \reviewstrike{on} a quantitative analysis of \setlength{\fboxsep}{0pt}\colorbox{cla}{a number of} publications in the NLP domain on the \setlength{\fboxsep}{0pt}\colorbox{flu}{collection}\reviewstrike{collecting}, publishing, and availability of research data. We find that\setlength{\fboxsep}{0pt}\colorbox{cla}{, although} a wide range of publications rely on data crawled from the web, \reviewstrike{but} few publications \setlength{\fboxsep}{0pt}\colorbox{style}{provide}\reviewstrike{give} details \setlength{\fboxsep}{0pt}\colorbox{gec}{of}\reviewstrike{on} how potentially sensitive data was treated. \setlength{\fboxsep}{0pt}\colorbox{redund}{In addition} \reviewstrike{Additionally, we find that}, while links to repositories of data are given, they often do not work\setlength{\fboxsep}{0pt}\colorbox{red}{,} even a short time after publication. We \setlength{\fboxsep}{0pt}\colorbox{style}{present}\reviewstrike{put together} several suggestions on how to improve this situation based on publications from the NLP domain, \setlength{\fboxsep}{0pt}\colorbox{style}{as well as} \reviewstrike{but} also other research areas. \\
\bottomrule
\end{tabular}
\end{center}
\caption{Examples of revision. Each edit type is highlighted respectively.}\label{tab:examples_revision} 
\end{table*}

We formalize the automated document revision task as follows. 
Given a source document $d$ that consists of paragraphs, an (possibly automated) editor $f$ revises $d$ into $d'$ ($f: d \mapsto d'$).
The revision $R$ is a set of edits $e$, and an edit $e$ is defined as a tuple $e=($\src, \tgt, $t$, $c)$, where \src~is the source phrase before the revision, \tgt~is the revised phrase, $t$ is the edit type (e.g., grammar, word choice, consistency), and $c$ is (optional) rationale comments to the edit. When \src~is empty (\O), this edit indicates \textit{insertion}, while it indicates \textit{deletion} in the case when \tgt~is empty. Otherwise, the edit is regarded as \textit{substitution}.
Automated document revision includes various edit types ($t$) such as mechanics, word choice, conciseness, and coherence. More details are discussed in \S\ref{subsec:edit_analysis}.
We emphasize that $t$ does not exclude the scope of conventional (sentential and subsentential) grammatical error and fluency correction.
Rationale comments ($c$) are useful resource for feedback generation study which becomes prominent in the GEC community~\cite{nagata-2019-toward,hanawa-etal-2021-exploring,nagata-etal-2021-shared}.
Automated document revision is thus a natural extension of sentence-level error correction to document-level error correction with a wider context being considered.
We will discuss the evaluation in \S\ref{sec:meta_eval_framwork}.

\section{The \corpus~Corpus}
\label{sec:dataset}

\subsection{Data Source}
\label{sec:data_source}
We use the ACL Anthology\footnote{\url{https://aclanthology.org}} as the data source of~\corpus~for the following reasons. 
First, we focus on \textit{document-level} revision rather than sentence-level revision, and therefore we select documents that have as few grammatical errors (i.e., conventional scope of GEC) as possible. 
The ACL anthology consists of peer-reviewed papers on NLP, and they are generally well written\footnote{In fact, one of the first shared tasks for grammatical error correction \cite{HOO2011} used the ACL Anthology as the source data, although the scope of annotation is limited to closed-class grammatical errors such as prepositions and determiners.}. 
Second, the ACL anthology contains a diverse range of papers in terms of authors and venues such as conferences vs. workshops, students vs. non-students, native vs. non-native speakers of English, as shown in \citet{bergsma-etal-2012-stylometric}.
Finally, the license and copyright of the ACL anthology are more flexible than existing datasets for similar purposes~\cite{lee-webster-2012-corpus}. 
We emphasize that a less restricted and widely accessible corpus would advance the research on automated document revision in the community.

We chose source documents from the ACL anthology as follows.
First, we created 8 (=$2^3$) groups based on the possible combinations of three different attributes: (1) whether the paper is published at a conference vs a workshop, (2) whether the paper is affiliated with a native vs non-native English speaking country, and (3) whether the first author is a student or not.
We randomly sampled papers until we have eight different papers per each group (i.e., 64 papers in total). 
For each paper, we extracted the title, the abstract, and the introduction as the source document ($d$) of ~\corpus.

\begin{table*}[t]
\centering
\small
\begin{tabular}{l || ccc || g} \toprule
                & \small{\citet{lee-webster-2012-corpus}} & \small{\citet{zhang-etal-2017-corpus}}     & \small{\citet{kashefi2022argrewrite}}    & \small{Ours (\corpus)}           \\ \midrule
\# docs         & 3760            & 60                & 86                & 64                \\
\# references &          1       &           1        &         1          &          3         \\
\% beyondGECs   & 3.2             & 49.4              & 52.6              & 56.9              \\
Drafted by      & ESL      & ESL/Native & ESL/Native & ESL/Native \\
Revised by      & Author          & Author            & Author            & Experts           \\
Feedback by     & Non-experts     & Experts           & Experts           & Experts           \\
Topic diversity &      \checkmark           &                   &                   &       \checkmark            \\
Public availability          &                 &           \checkmark        &       \checkmark            &      \checkmark          \\   \bottomrule   
\end{tabular}
\caption{Characteristics of \corpus~corpus compared to existing document revision corpus. \% beyondGECs shows the ratio of edits that are not covered by GEC edit types. \textit{Drafted by} indicates who wrote the (first) draft, \textit{Revised by} shows who revised the draft by whose feedback (\textit{feedback by}). Topic diversity (\checkmark) presents whether the corpus contains two or more topics, or just single topic (no \checkmark). Public availability (\checkmark) shows whether the corpus is publicly available to the community.}
\label{tab:statistics}
\end{table*}

\subsection{Annotation Scheme}
The scope and granularity of edit types has also a wide variety in previous work and there is no standard set of labels. 
Thus, we define edit type categories (Table~\ref{tab:edit_type}) based on previous literature on the argumentative and discourse writing~\cite{Kneupper,faigley1981analyzing,burstein2003finding,zhang-etal-2017-corpus}.
Table~\ref{tab:examples_revision} shows concrete examples of each edit type in~\corpus.

In terms of the format of the annotation, there are several annotation schemas used for document-level revision and grammatical error correction.
For example, \citet{HOO2011}, \citet{HOO2012}, and \citet{Yannakoudakis:11:ACL} adopt an XML format, \citet{zhang-etal-2016-argrewrite} uses a simple table format, \citet{berzak-etal-2016-universal} uses a CoNLL format~\cite{buchholz-marsi-2006-conll}, and \citet{dahlmeier:2012:M2} have created an M2 format which is a variant of the CoNLL format.

For creating~\corpus, we choose an XML format for the following reasons.
First, XML is easy to parse with standard libraries (e.g., Python ElementTree, Java DOM parser)\footnote{We made the nest of XML tags as shallow as possible for users to parse documents even more easily. In~\corpus, the maximum depth of nested XML tags is two.} compared with the other formats that often require exclusive scripts. These exclusive scripts will cost more for the maintenance to keep up with the updates of additional dependency.
Second, XML is more flexible than the other formats to embed additional information such as edit types, edit rationale, comments, and other meta information.
Example of our XML annotation is shown in Appendix (Table~\ref{tab:xml}).

\subsection{Annotators}
We recruited three professional editors who are native speakers of English and also have years of experience in editing and proofreading English academic writing.
The editors independently revised all the 64 documents on Google Docs platform, adding edit rationale whenever appropriate.
The revised documents are converted into the XML format by the first two authors.\footnote{While converting, we also made minor corrections and remapping edit types only when it is necessary.}

\subsection{Statistical Analysis}
\label{subsec:edit_analysis}

The right-most column in Table~\ref{tab:definition} shows the distribution of edit types found in randomly sampled 16 papers (25\% of \corpus~corpus).
We find that 56.9\% of the edits are related to the scope beyond sentence-level context (e.g., redundancy), which is larger than the other document revision corpora (Table~\ref{tab:statistics}). 
This is simply because \corpus's source documents are academic papers which are already proofread to some extent, compared with other existing document revision corpora where language learner essays are used as the source.
It is also important to note that paragraph-level revisions have longer range (i.e., more tokens in \textit{one edit}) and thus have fewer numbers than word- and sentence-level edits in its nature.
In terms of the difference according to three different attributes (\S~\ref{sec:data_source}), we did not find any clear patterns, indicating that the quality of papers in the ACL corpus is uniformly good across venues and author attributes. 
The details are shown in Appendix (Table~\ref{tab:attributes}).

In document-level revision, it is not straightforward to compute inter-annotator agreement because of the diversity of possible revisions and wide spans of edits.
Thus, we measured two levels of inter-annotator agreements (1) agreement on \textit{detection} and (2) agreement on \textit{correction}.
The first measurement computes how often edit spans overlap (i.e., agree) among annotators, and the second measurement  computes how often edit labels match when two or more annotators detect the same (or overlapped) span.
Table~\ref{tab:agreement} shows the results. 
The agreement rate on \textit{detection} is about 0.3, indicating the diversity of possible revisions, while the agreement on \textit{correction} is about 0.8, which shows that edits are consistent among annotators when they  correct the same (local) span.
\begin{table}[t]
\centering
\begin{tabular}{llll}\toprule
    Levels           & Avg  & Min  & Max  \\ \midrule
Detection    & 0.32 & 0.27 & 0.35 \\
Correction & 0.83 & 0.75 & 1.00 \\ \bottomrule
\end{tabular}
\caption{Two levels of inter-annotator agreements: agreement on \textit{detection} and \textit{correction}. We find that annotators find out diverse possible revision while they make similar (or the same) revisions if they edit the same local span.}
\label{tab:agreement}
\end{table}
\begin{figure*}[t]
 \centering
  \includegraphics[width=1.0\linewidth]{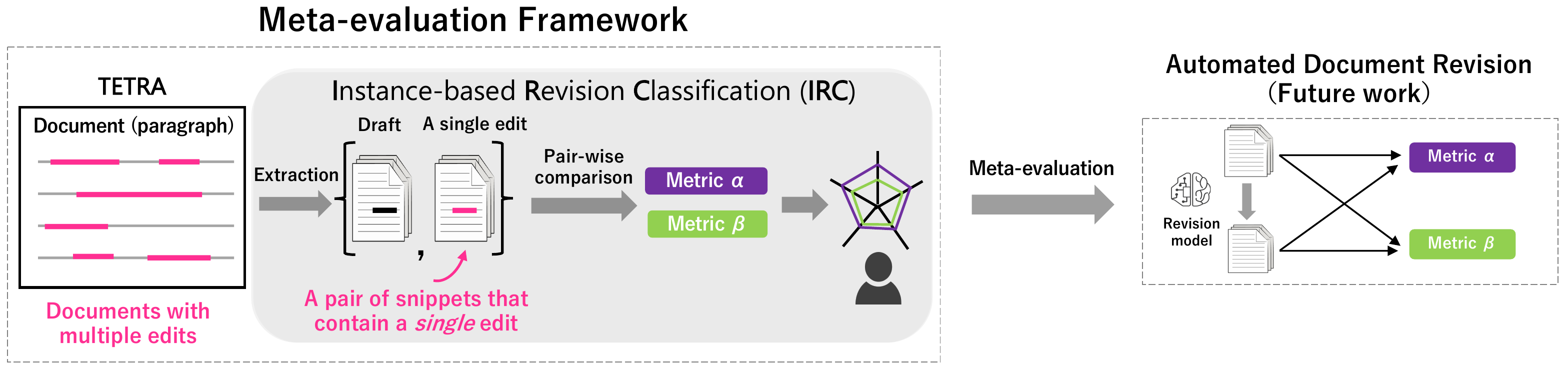}
 \caption{Overview of proposed meta-evaluation framework. We introduce document revision corpus (\corpus) and propose instance-based revision classification~\meta~to measure (i.e., meta-evaluate) the improvement of documents by the revision.}
 \label{fig:metaeval_overview}
\end{figure*}

\section{\meta: Meta-evaluation Framework}
\label{sec:meta_eval_framwork}
Towards automated document revision, in addition to creating a corpus, it is essential to establish evaluation metrics that measures document's quality improvement (and possibly deterioration) by revision.
However, it is not feasible to elicit all possible gold references because there are infinite ways of document revision.
Moreover, it is difficult to automatically measure the quality of a revision based on an \textit{absolute} metric because a single document contains a variety of edits based on many evaluation aspects (Table \ref{tab:definition}).

Thus, it is more straightforward to consider \textit{relative} metric which, given a pair of documents, detects if one is an improved version of the other based on binary classification.
Pairwise comparisons have been shown to be effective as a meta-evaluation method in situations where absolute evaluation is difficult in previous studies~\cite{guzman-etal-2015-pairwise,NIPS2017_d5e2c0ad}.
However, in the document revision scenario, it is still challenging to perform make relative judgements because one lumps together a variety of edits and the binary value would not tell which of the edit(s) exactly contribute the improvement.
In fact, there is a limit in the comprehensive evaluation, as the optimal metric varies depending on the evaluation aspects~\cite{kasai2021bidimensional,kasai2021transparent}.

Addressing the above concerns, we propose \textbf{i}nstance-based \textbf{r}evision \textbf{c}lassification (\meta), which can detect quality improvements by document revision.
In \meta, more concretely, given a pair of snippets that contain a \textit{single} edit, we compare (reference-less) metrics according to the accuracy of binary prediction (i.e., which of the snippets is a revised one?).
By focusing on comparing `single edit' differences, we can obtain transparent and interpretable measures for each edit type (e.g., which edit type is more challenging to revise than the other types), which enables us to investigate better evaluation metrics in future. 
Figure~\ref{fig:metaeval_overview} shows an overview of \meta~framework.

\section{Experiment}
\label{sec:experiment}
In this section, we demonstrate the utility of \meta~framework by evaluating baseline metrics.
Specifically, we compare BERT~\cite{devlin-etal-2019-bert} and GPT-2~\cite{radford2019language} as supervised and unsupervised baseline metrics to see whether pre-trained language models can be a good baseline metric to discriminate the original and revised snippets. 
In addition, we investigate the current status and feasibility of automatic evaluation for document-level revisions.

\subsection{Evaluation}
To conduct meta-evaluation with \meta, we need to convert \corpus~corpus into pairs of snippets that contain a single edit.
Thus, we divided \corpus~corpus into 3 (48 papers):1 (16 papers) for train set:test set in terms of paper units, and then converted the test data into pairs of snippets.
If multiple edit type was assigned, each edit type was extracted independently as a snippet pair.
Furthermore, the context width when extracted as snippets was in paragraphs, assuming a single paragraph to be a single document.
Using the above procedure, we obtained 1,368 snippet pairs for evaluation.

\subsection{Baseline metrics}
We employed the following two reference-less (unsupervised and supervised) metrics with pre-trained neural language models as baselines, using the Pytorch implementation of ~\texttt{transformers}~\cite{wolf-etal-2020-transformers}.

\paragraph{GPT-2 based metric (\gpt)}
It is an unsupervised evaluation metric using GPT-2~\cite{radford2019language}.
This metric compares the per-word perplexity of each of the two input documents. It performs a binary prediction based on the hypothesis that if the per-word perplexity of the revised document is lower than that of the source document, the revised one is a good revision, and vice versa.

\paragraph{BERT based metric (\bert)}
It is a supervised evaluation metric based on binary prediction using BERT~\cite{devlin-etal-2019-bert}.
For fine-tuning, we used the train split of \corpus~(resulting in 868 document pairs) with half of the document pairs randomly swapped before and after revisions for creating negative examples (deterioration version) and fine-tuned on them as a binary classification problem task.
The hyperparameters for the model training are shown in Appendix~\ref{appendix:hyper_parameter} (Table~\ref{tab:hyperparameters}).

\subsection{Results}
\begin{figure}[t]
 \centering
  \includegraphics[width=0.9\linewidth]{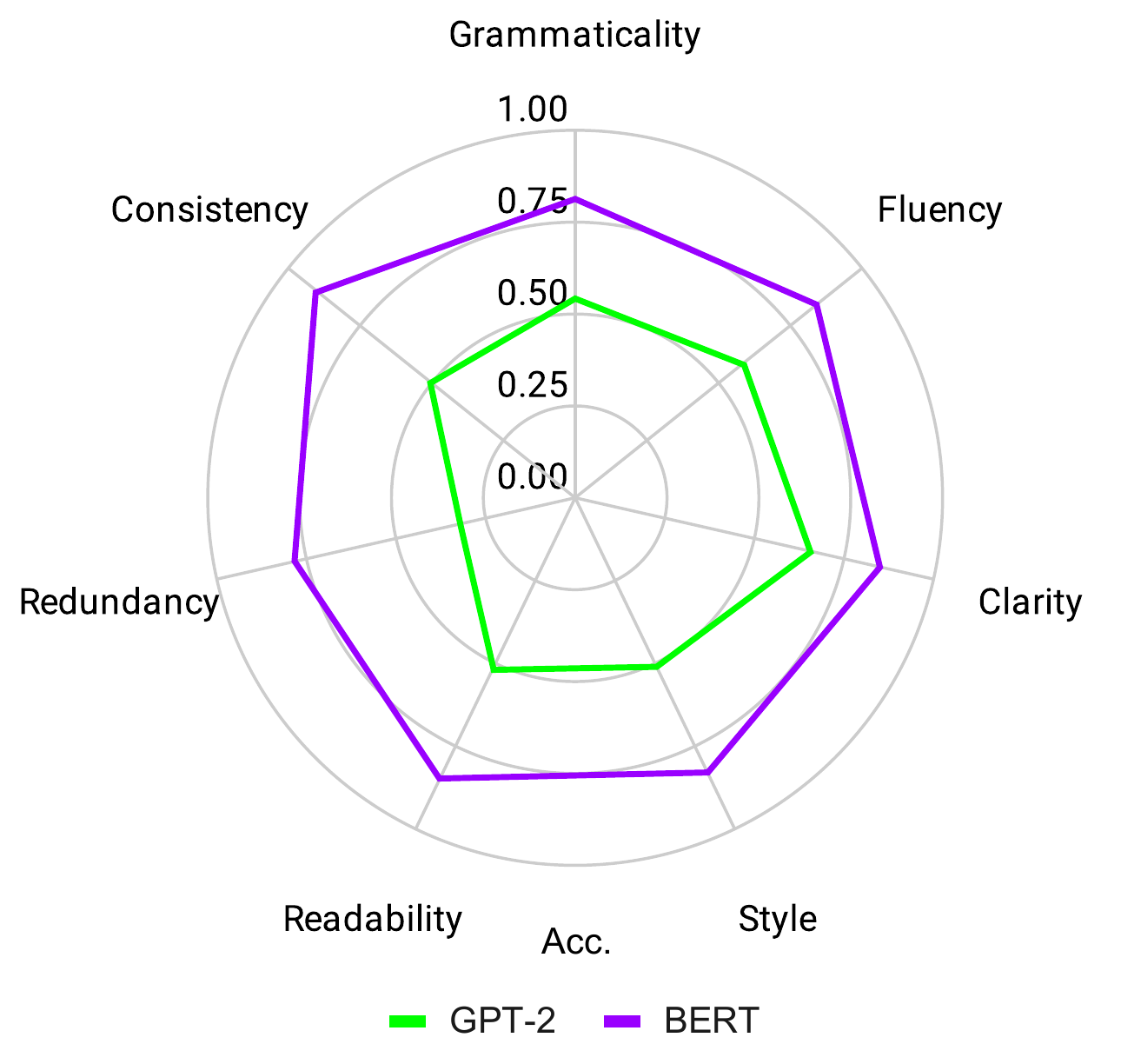}
 \caption{Meta-evaluation result (Accuracy).}
 \label{fig:aspect}
\end{figure}

As shown in Figure~\ref{fig:aspect}, our \meta~framework enables us to evaluate the accuracy of each metrics in aspect-wise format and to focus on developing optimal evaluation metrics for each evaluation aspect while analyzing the strengths and weaknesses of them.
We also find that the supervised metric (\bert) is able to classify with 0.79 - 0.90 accuracy, indicating the supervised metric based on pre-trained neural language models can be a good baseline metric to discriminate the original and revised snippets even when difference is subtle.

\section{Analysis}

\subsection{Is \meta~framework reliable?}

The experiment in \S\ref{sec:experiment} has shown the utility of \meta~framework, but its reliability is not clear.
In other words, the following questions are naturally raised: \textit{do improved accuracy metrics on \meta~framework mean that they can more accurately determine whether a revision is good or bad?}
For example, it remains possible that the supervised based metric (\bert), trained on the binary classification of a source document and its revisions by the experts, is not judging whether the revision is good or bad but whether it is expert or not by finding expert-specific phrases and expressions.

To verify the above question and confirm the reliability of \meta~framework, we evaluate the performance of the baseline metrics by introducing ``worse-quality revisions'' into \corpus~by using corruption methods that artificially worsen the quality of the source documents.
Suppose the performance of the metrics is significantly degraded by introducing the worse-quality revisions not included in the training data.
In that case, it is more likely that the metrics do not judge whether a revision is good or bad but whether it is expert or not, and vice versa.

\subsubsection{Corruption Methods}
We employed the following two simple corruption methods.

\paragraph{Automatic Error Generation (AEG): } 
It is a method to artificially generate grammatical errors for training data augmentation in GEC that has been actively studied in recent years.
We used a back-translation model, the most commonly used in GEC among the AEGs~\cite{Xie:18:NAACL,kiyono-etal-2019-empirical,koyama-etal-2021-comparison}, with the motivation to generate worse-quality revisions mainly in terms of \textit{grammaticality} and \textit{fluency}.
A reverse model, which generates an ungrammatical sentence from a given grammatical sentence, is trained in the back-translation model.
As for building the reverse model, we followed the general settings in previous studies~\citep{kiyono-etal-2019-empirical,koyama-etal-2021-comparison}.
The detailed of experimental setting for our AEG model is shown in Appendix~\ref{appendix:setupaeg}

\paragraph{Sentence Shuffling}
It is a simple corruption method that deteriorates the quality of a source document by randomly shuffling sentences.
As illustrated in Figure~\ref{fig:overview}, document revision involves reordering sentences in terms of \textit{flow} and \textit{consistency} of argumentation.
In this study, we applied sentence shuffling to the true distribution of \textit{consistency}, i.e., 5\% of documents (See Table~\ref{tab:definition}) with worse-quality revisions generated by the AEG.

\subsubsection{Result}

\begin{table}[t]
\centering
\begin{tabular}{lrr}
\toprule
Test sets & \bert & \gpt \\ \midrule
\corpus~$_{\{source, gold\}}$             & 0.85 & 0.57  \\
worse only$_{\{source, worse\}}$                & 0.96 & 0.81  \\ \bottomrule
\end{tabular} 
\caption{Evaluation results for reliability.}\label{tab:exp_pseudo}
\end{table}

Table~\ref{tab:exp_pseudo} shows the performance of the baseline metrics on the test set consisting of source and gold revisions (\corpus) and on the test set consisting of source and worse-quality revisions (worse only).
We find that \bert~on the worse only was further improved by points, which indicates that the fine-tuned pre-trained language model can discriminate the quality of documents after revision.
This also shows the reliability of \meta~framework based on the binary classification of source documents and its gold revisions, which does not include explicit worse-quality revisions.

\subsection{Do existing metrics not work?}
\begin{table}[t]
\centering
\begin{tabular}{lrr}
\toprule
Outputs                & ERRANT & GLEU  \\ \midrule
Source       & 0.0 (0.0)    & 70.6 (1.5)\\ 
Human experts          & 24.5 (5.7)  & 71.4 (1.0)\\ \midrule
\end{tabular}
\caption{Evaluation results with GEC's metrics.Values in parentheses indicate standard deviations.}\label{tab:gec_metric}
\end{table}
A motivation of this study is based on the assumption that commonly used reference-based metrics in GEC cannot accurately evaluate document revisions.
To verify this assumption, we evaluate gold revisions by human experts on \corpus~using existing GEC's metrics and analyze whether the existing metrics does not work in document revisions.

\subsubsection{Examined metrics}
\label{subsec:metrics}
We use ERRANT~\cite{bryant:2017:automatic} and GLEU~\cite{napoles:2016:gleu}, which are widely used in GEC, as the evaluation metrics to be validated.
The details of each are described below.

\paragraph{ERRANT}
It is an improved version of the previously standard metric, Max Match (M$^2$) Scorer~\cite{dahlmeier:2012:M2}.
Similar to M$^2$ Scorer, ERRANT is performed based on the Max Match method, which identifies the maximum match using the edit lattice when matching the edits between systems and references, but the method of edit extraction differs from that of M$^2$ Scorer.

\paragraph{GLEU}
It is an improved version of BLEU~\cite{papineni-EtAl:2002:ACL}, the most \textit{de facto} evaluation metric in machine translation, for GEC.
GLEU is computed by subtracting the number of n-grams that appear in the input but not in the reference from the number of n-grams that match in the system output and reference.
It is known to be more highly correlated with human judgment than M$^2$ Scorer~\cite{Napoles2016}.

\subsubsection{Result}
Table~\ref{tab:gec_metric} shows the evaluation result.
Note that since three gold revisions by human experts are assigned to \corpus, the values represent the average of the three.
The evaluation results with ERRANT show that even human experts have a low value of 24.5 points, implying that it has issues in evaluating document revisions.
ERRANT evaluates systems based on the extent to which the edit span suggested by systems matches the gold edit span included in references.
However, in document revisions that require cross-sentence editing or more dynamic editing, ERRANT may have difficulty extracting accurate edit spans and matching them with references.

On the other hand, GLEU may seem to work as an evaluation metric since it succeeds in giving somewhat higher scores to human experts' revisions.
However, GLEU also has issues, since its evaluation score for source documents, i.e., outputs that without any editing, are comparable to those of human experts.
GLEU score is basically computed based on the n-gram agreement ratio in the three sentences (documents in this case): input, system output, and reference.
In document revision, a task with low agreement rates \S\ref{subsec:edit_analysis}), GLEU, which performs document-by-document matching, suggests that it tends to overestimate unedited output.

\section{Conclusion}
To go beyond \textit{sentence-level} automated grammatical error correction to \textit{document-level} revision, we proposed the new task of automated document revision and also provided the new corpus (\corpus) and the meta-evaluation framework (\meta), which allows to run transparent and interpretable analysis for better designing evaluation metrics.
In addition, we explored reference-less and interpretable methods that can detect quality improvements by document revision.
Our experimental results show that a fine-tuned pre-trained language model can discriminate the quality of documents after revision even when difference is subtle, indicating the feasibility of automated document revision.
These research foundations and promising result will encourage the community to further study automated document revision models and metrics beyond sentence-level error corrections.

\bibliography{emnlp2022}
\bibliographystyle{acl_natbib}

\newpage
\appendix

\section{Example of XML annotation}
\label{appendix:sample_xml}
\begin{table*}[t]
\small
\begin{tabular}{p{16cm}}
\toprule
\begin{lstlisting}[language=xml]
<doc id="Pxx-xxxx" editor="A" format="Conference" position="Non-student" region="Native">
<abstract>
<text>In this paper, (...) extracted sense inventory. The</text>
<edit type="conciseness" crr="induction and disambiguation steps" comments="conciseness - just tightening it up a little bit.">induction step and the disambiguation step</edit>
<text>are based on the same principle: (...) topical dimensions</text>
<edit type="readability" crr=". In" comments="readability - this sentence is getting a bit long, so splitting it in two here.">; in</edit>
<text>a similar vein, ...</text>
...
</abstract> 
<introduction>
<text>Word sense induction (...)</text>
<text>\n\n Word sense disambiguation (...)</text>
<edit type="punctuation" crr="" comments="punctuation - comma is not appropriate.">,</edit>
...
</introduction>
\end{lstlisting} 
\\
\bottomrule
\end{tabular}
\caption{Example of XML annotation. For brevity, we omitted a part of the text with ``...''. }
\label{tab:xml}
\end{table*}

\section{Aspect distribution}
\label{appendix:aspect_dist}

\begin{table*}[t]
\centering
\small
\begin{tabular}{l  rgrg  rgrg  rgrg}\toprule
  & \multicolumn{2}{c}{\begin{tabular}{c} Student \end{tabular}} & \multicolumn{2}{c}{\begin{tabular}{c} Non-student \end{tabular}} & \multicolumn{2}{c}{\begin{tabular}{c} Native \end{tabular}} & \multicolumn{2}{c}{\begin{tabular}{c} Non-native \end{tabular}} 
& \multicolumn{2}{c}{\begin{tabular}{c} Conf. \end{tabular}} & \multicolumn{2}{c}{\begin{tabular}{c} WS \end{tabular}} \\
\cmidrule(r){2-5}\cmidrule(r){6-9}\cmidrule(r){10-13}
\multicolumn{1}{l}{Aspects}  & \#  & \%   &  \#   & \% & \#  & \%   &  \#   & \% & \#  & \%   &  \#   & \% \\
\midrule
Grammaticality & 79 & 19.5 & 106 & 21.5 & 60  & 16.5  & 125  & 21.3  & 110  & 22.7 & 75  & 16.2       \\
Fluency & 115  & 25.2 & 110   & 22.4   & 74  & 20.4 & 151  & 25.8 & 99   & 20.4 & 126  & 27       \\
Clarity  & 100  & 21.9 & 84 & 17.1  & 88 & 24.2  & 96  & 16.4 & 84  & 17.3  & 100  & 21.6       \\
Style  & 39 & 8.5  & 37 & 7.5   & 29 & 8.0  & 47  & 8.0   & 46  & 9.5   & 30 & 6.5        \\
Readability  & 74   & 16.2 & 85 & 17.3  & 75 & 20.7 & 84 & 14.3  & 92  & 19.0   & 67 & 14.4  \\
Redundancy  & 32 & 7.0  & 36 & 7.3   & 22   & 6.1  & 46 & 7.8   &  25   & 5.2  & 43  & 9.3  \\
Consistency & 18 & 3.9  & 34 & 6.9   & 15   & 4.1  & 37 & 6.3   & 29 & 6.0   & 23  & 5.0     \\ \bottomrule 
\end{tabular}
\caption{Distributions of revision aspects by writer's attributes.}
\label{tab:attributes}
\end{table*}

\section{Hyper-parameters settings}
\label{appendix:hyper_parameter}

\begin{table}[t]
\centering
\begin{tabular}{@{}lp{40mm}@{}}
\toprule
 Configurations & Values \\ \midrule
 Model Architecture & bert-base-uncased \\
 Optimizer& Adam~\cite{kingma:2015:ICLR}\\
 Learning Rate &  2e-5\\
 Number of Epochs & 10 \\
 Batch Size & 32 \\
 \bottomrule
\end{tabular}
\caption{Hyper-parameters settings}\label{tab:hyperparameters}
\end{table}

\section{Experimental settings for AEG}
\label{appendix:setupaeg}
We adopted the ``Transformer (big)'' settings~\citep{Vaswani:17:NIPS} using the implementation in the \texttt{fairseq} toolkit~\citep{ott2019:arxiv:fairseq} as a revise GEC model.
In addition, we used the BEA-2019 workshop official dataset~\cite{bryant-etal-2019-bea} as the training and validation data.
For preprocessing, we tokenized the training data using the \texttt{spaCy} tokenizer\footnote{\url{https://spacy.io/}}.
Then, we removed sentence pairs where both sentences where identical or both longer than 80 tokens.
Finally, we acquired subwords from the target sentence via the byte-pair-encoding (BPE)~\citep{sennrich:2016:ACL} algorithm.
We used the \texttt{subword-nmt} implementation\footnote{\url{https://github.com/rsennrich/subword-nmt}} and then applied BPE to splitting both source and target texts.
The number of merge operations was set to 8,000.

\end{document}